\documentclass[journal]{vgtc}                
\ifpdf
  \pdfoutput=1\relax                   
  \usepackage{graphicx}                
  \DeclareGraphicsExtensions{.pdf,.png,.jpg,.jpeg} 
\else
  \usepackage{graphicx}                
  \DeclareGraphicsExtensions{.eps}     
\fi%

\graphicspath{{figures/}{pictures/}{images/}{./}} 

\usepackage{microtype}                 
\PassOptionsToPackage{warn}{textcomp}  
\usepackage{textcomp}                  
\usepackage{mathptmx}                  
\usepackage{times}                     
\usepackage{cite}                      
\usepackage{tabu}                      
\usepackage{booktabs}                  
\usepackage{amsfonts}
\usepackage{amsmath}
\usepackage{multirow}

\vgtccategory{Research}
\vgtcpapertype{algorithm}

\title{An Interpretable Neuron Embedding for Static Knowledge Distillation}

\author{Wei Han, Yangqiming Wang, Christian Böhm, Junming Shao}
\authorfooter{
\item
 Wei Han and Yangqiming Wang are with University of Rlectronic Science and Technology of China. E-mail: \{weihan, leo\_wang\}@std.uestc.edu.cn.
\item
 Christian Böhm is with Ludwig Maximilian Muenchen Unitversitaet. E-mail: boehm@ifi.lmu.de.
\item
 Junming Shao is with University of Rlectronic Science and Technology of China. E-mail: junmshao@uestc.edu.cn.
}

\abstract{Although deep neural networks have shown well-performance in various tasks, the poor interpretability of the models is always criticized. In the paper, we propose a new interpretable neural network method, by embedding neurons into the semantic space to extract their intrinsic global semantics. In contrast to previous methods that probe latent knowledge inside the model, the proposed semantic vector externalizes the latent knowledge to static knowledge, which is easy to exploit. Specifically, we assume that neurons with similar activation are of similar semantic information. Afterwards, semantic vectors are optimized by continuously aligning activation similarity and semantic vector similarity during the training of the neural network. The visualization of semantic vectors allows for a qualitative explanation of the neural network. Moreover, we assess the static knowledge quantitatively by knowledge distillation tasks. Empirical experiments of visualization show that semantic vectors describe neuron activation semantics well. Without the sample-by-sample guidance from the teacher model, static knowledge distillation exhibit comparable or even superior performance with existing relation-based knowledge distillation methods.} 

\keywords{Neural network interpretability, semantic embedding, knowledge distillation}

\CCScatlist{ 
 \CCScat{K.6.1}{Management of Computing and Information Systems}%
{Project and People Management}{Life Cycle};
 \CCScat{K.7.m}{The Computing Profession}{Miscellaneous}{Ethics}
}

\vgtcinsertpkg

\begin{document}


\firstsection{Introduction}

\maketitle

As high-performance neural network models are gradually applied to real scenarios, there is still a trust gap between human beings and artificial intelligence. It is hard to completely hand over ourselves to a non-interpretable model. In order to open the black box of neural networks, the research of neural network interpretability has received increasing attention. Existing methods mainly focus on analyzing the prediction results of a trained neural network model. For instance, given similar examples of the current sample~\cite{caruana1999case,bien2011prototype}, point out the important attribute in the input sample~\cite{shapley2016games,zhou2016learning}, perturbation analysis for the neurons plays an important role in the specified task~\cite{fong2017interpretable,zintgraf2017visualizing}, etc. However, a single misjudgment in areas with high impact outcomes such as autonomous driving, financial prediction, and medical diagnosis can mean a major accident. Even if we can trace back the input that triggered the miscalculation afterwards or activated neurons at the time of the wrong prediction, it is difficult to recover from the accident that has already occurred. Therefore, how to give a sample- and class-independent explanation of all neurons in a model before the model is put into practice is a key issue to enhance the trustworthiness of neural network models.

Existing interpretable methods do not provide a comprehensive explanation of neurons. Making interpretations for the neurons corresponds to semantic-based interpretation models. The classic method, NetDissect~\cite{bau2017network}, matches the activation of a neuron to an artificially given semantic annotation in order to give a semantic interpretation of the neuron. The experimental results show that one-third of the neurons do not have a specific semantic meaning, while one neuron may correspond to more than one semantic meaning. It is probably due to the fact that many hidden patterns are not given semantics and there is a correlation between semantics. In order to describe a neuron completely, we propose to embed the neuron as a semantic vector, rather than assigning a single semantic tag. Once a neuron has a materialized semantic vector, the knowledge of the neuron is extracted. Therefore, the semantic relationship between different neurons is capable to be intuitively compared, analyzed, and further exploited for many interesting applications. For instance, can we use this extracted semantic relationship as the static knowledge to instruct a student model for knowledge transfer?

\begin{figure}[t]
	\centering
	\includegraphics[width=8cm,height=5.3cm]{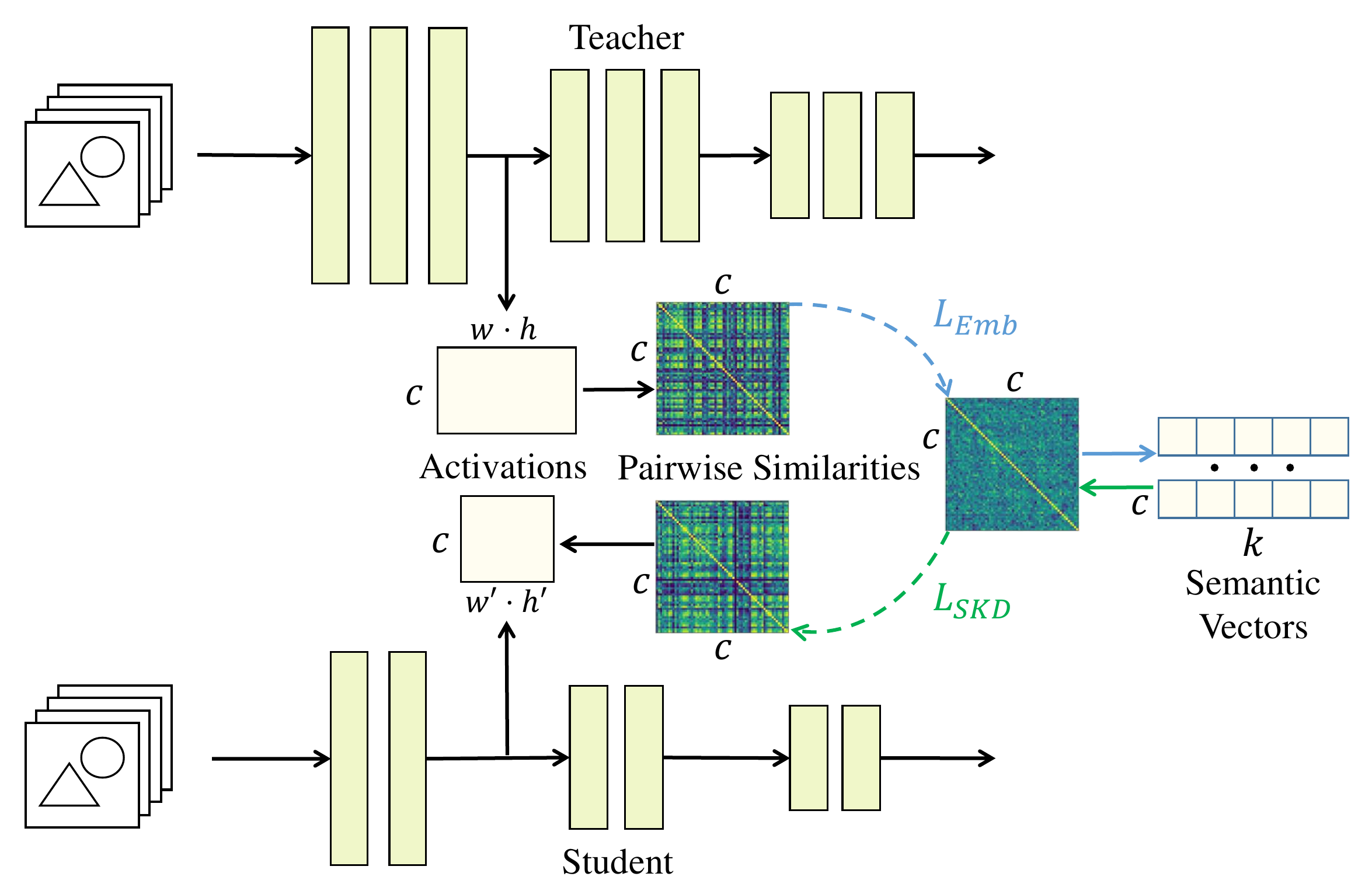}   
	\caption{Schematic diagram about neuron embedding and static knowledge distillation. During semantic embedding, we assume that neurons with similar activation have similar semantics, and vice versa in knowledge transfer. By aligning activation similarity and semantic vector similarity in two independent phases, we achieve the embedding of neuron semantics and knowledge distillation. The knowledge obtained through semantic embedding is static and is not dependent on individual instances or classes. It allows for an intuitive analysis and exploitation of neuron global semantics. As the two steps in knowledge distillation are capable to be decoupled, the knowledge transfer is achieved without sample-by-sample guidance.} 
	\label{fig:model}
\end{figure}

\begin{figure*}[t]
	\centering
	\includegraphics[width=16.2cm,height=7cm]{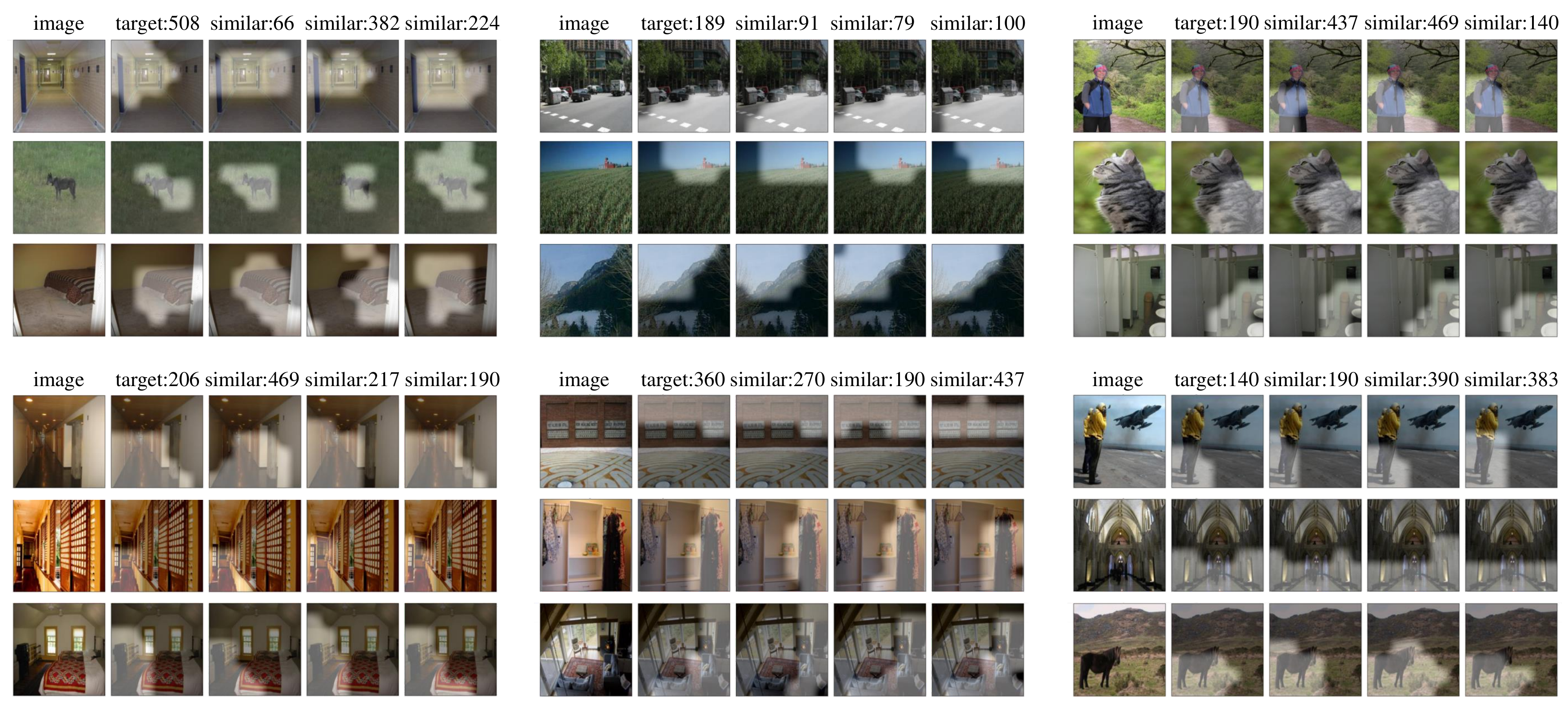}   
	\caption{Visualization of neuron similar relation. To show the alignment between semantic vector similarity and activation map similarity, we demonstrate the activation maps of top-layer neurons and their three most similar neurons in different images. The similarity between neurons is directly measured by the optimized semantic vectors. The ID of neurons is given. The activation area of neurons is covered by a light mask.} 
	\label{fig:relationship}
\end{figure*}

Traditional knowledge distillation approaches~\cite{hinton2015distilling,yim2017gift,romero2014fitnets} employ a pre-trained high-performance teacher model to give the guideline of each sample in the student model training. As two-stage distillation requires high computation and storage costs, it is not efficient and not scalable. Although teacher-free knowledge distillation methods have been proposed, they almost perform as self-distillation~\cite{furlanello2018born,crowley2018moonshine,zhang2019your}. By focusing on the re-integration of information within the student model or between peers, such methods do not enable directed knowledge transfer across models. In contrast, The semantic vector of neurons is regarded as the neuron-intrinsic global knowledge, which is not dependent on individual samples or task classes. Such static knowledge is potential information to be employed to directly instruct the student model without sample-by-sample guidance from the teacher model. 

In this paper, we propose to encode neurons into the semantic space to extract a neuron-intrinsic task-independent global interpretation. Specifically, we assign each neuron in the convolutional layer a unique semantic vector. Neurons with similar activation for the same instance are assumed to present similar semantics. The semantic vectors of the neurons are optimized by continuously aligning semantics with activation in stochastic gradient iterations. Based on semantic vectors, the relationship between neurons can be explored intuitively by comparing the similarity. Besides demonstrating the direct similarity between neuron knowledge, we also investigate the additive relation between neurons. Furthermore, we introduce the relationship between neurons by semantic vectors as static knowledge to guide student models, achieving a knowledge distillation without sample-by-sample guidance from the teacher model. Ultimately, various visualizations qualitatively demonstrate the plausibility of the interpretability given by neural semantic vectors, while experiments on knowledge distillation across multiple architectures quantitatively show the validity of the interpretability.

\section{Methodology}
\label{sec:method}

In this section, we demonstrate how to embed the knowledge inside neurons into the semantic space, and how to achieve a static knowledge distillation without sample-by-sample guidances from the teacher model.

\subsection{Neuron Embedding}


We embed neurons into the semantic space in the form of vectors, to achieve a comprehensive representation of neuron semantics and the potential to conveniently exploit global semantics. The top half of Figure~\ref{fig:model} visualizes the procedure of neuron embedding. Specifically, for a pre-trained neural network model, we propose to abstract the semantic information of each neuron into a semantic vector. Given a specific semantic vector of the neuron in each layer, we assume that neurons with similar activation maps for the same image should have similar semantics, similar to Word2vec~\cite{mikolov2013distributed}. Therefore, the semantic representations of neurons are optimized by continuously aligning activation map similarity and semantic vector similarity during training. For the neurons in the $l$ layer, we randomly initialize their semantic vectors $S^l \in \mathbb{R}^{C \times K}$, where $C$ is the number of neurons in the $l$ layer and $K$ is the dimension of the semantic vector. During the forward processing of the neural network, neurons in the $l$ layer give their activation $A^l \in \mathbb{R}^{C \times WH}$. The similarity metric $d(\cdot)$ yields the pairwise activation similarity $d(a^l_i, a^l_j)$ and the pairwise semantic vector similarity $d(s^l_i, s^l_j)$, and the cosine similarity is used as the metric $d(\cdot)$ in this paper. Eventually, the embedding loss matches the two similarities in the framework of cross-entropy. The pairwise similarity value needs to be processed via the softmax function $h(\cdot)$ to transform into probability values in the cross-entropy loss.

\begin{equation}
\label{equ:embedding}
\begin{aligned}
	 L_{emb} &= - \frac{1}{L} \cdot \frac{1}{C} \sum_l^L \sum_{i,j}^C p(j|i) \log q(j|i) \\
	 			   &= - \frac{1}{L} \cdot \frac{1}{C} \sum_l^L \sum_{i,j}^C h(d(a^l_i, a^l_j)) \log h(d(s^l_i, s^l_j))
\end{aligned} 
\end{equation}

\begin{table*}[t]
\centering
\caption{Knowledge distillation experiments on CIFAR-100.}
\label{tab:kd}
\begin{tabular}{p{2.0cm}p{1.5cm}p{1.5cm}p{1.5cm}p{1.5cm}p{1.5cm}p{1.5cm}ccccc}
\toprule 
	&	Teacher 		& WRN-40-2 & resnet56	& resnet110	& resnet110 & resnet32x4 & vgg13\\
	&	Student 		& WRN-16-2 & resnet20	& resnet20	& resnet32  & resnet8x4  & vgg8\\
\cmidrule{2-8}	
	&	Teacher 		& 75.61 & 72.34	& 74.31	& 74.31  & 79.42 & 74.64	\\
	&	Student 		& 73.26 & 69.06	& 69.06	& 71.14  & 72.50 & 70.36	\\
\midrule

\multirow{2}{*}{Logit-based}
	&	KD 				& 74.92 & 70.66 & 70.67 & 73.08 & 73.33 & 72.98 \\
	&	PKT 			& 74.54 & 70.34 & 70.25 & 72.61 & 73.64 & 72.88 \\
	&	CRD 			& 75.48 & 71.16 & 71.46 & 73.48 & 75.51 & 73.94 \\
\midrule

\multirow{6}{*}{Feature-based}
	&	FitNet 			& 73.58 & 69.21 & 68.99 & 71.06 & 73.50 & 71.02 \\
	&	AT 				& 74.08 & 70.55 & 70.22 & 72.31 & 73.44 & 71.43 \\
	&	VID 			& 74.11 & 70.38 & 70.16 & 72.61 & 73.09 & 71.23 \\
	&	AB 				& 72.50 & 69.47 & 69.53 & 70.98 & 73.17 & 70.94 \\
	&	FT 				& 73.25 & 69.84 & 70.22 & 72.37 & 72.86 & 70.58 \\
	&	NST 			& 73.68 & 69.60 & 69.53 & 71.96 & 73.30 & 71.53 \\
\midrule
\multirow{5}{*}{Relation-based}
	&	FSP 			& 72.91 & 69.95 & 70.11 & 71.89 & 72.62 & 70.23 \\
	&	SP 				& 73.83 & 69.67 & 70.04 & 72.69 & 72.94 & 72.68 \\
	&	CC 				& 73.56 & 69.63 & 69.48 & 71.48 & 72.97 & 70.71 \\
	&	RKD 			& 73.35 & 69.61 & 69.25 & 71.82 & 71.90 & 71.48 \\
	&	\textbf{SKD(Ours)}	& 73.79 & 70.23 & 70.27 & 71.86 & 72.91 & 71.09 \\

\bottomrule
\end{tabular}
\end{table*}

\subsection{Static Knowledge Distillation}


As the extracted neuron semantic vector is the static knowledge of the neural network, we introduce the semantic relation-based information as a prior to realize the decoupled two-stage knowledge transfer. Compared with previous instance relation-based knowledge distillation methods, the proposed method avoids sample-by-sample guidance from the teacher model. There is only forward processing and backpropagation of the student model during knowledge distillation, as illustrated in the bottom half of Figure~\ref{fig:model}. It provides a promising way for efficient and scalable knowledge distillation. Specifically, in contrast to embedding loss, we assume that neurons that are semantically similar in the prior of the teacher model should have similar activations for samples. Static knowledge distillation employs the similarity between neurons as the target and the activation of the neural network as the variable. By the cosine similarity measure, the pairwise similarity matrix $d(S^L, S^L)$ of the semantic vectors and the pairwise similarity matrix $d(A^L, A^L)$ of the activation maps in the corresponding layer of the student model are calculated, respectively. Then, the softmax function $h(\cdot)$ probabilize them. In the form of cross-entropy, the similarity of neuron semantics and the similarity of neuron activation maps are re-aligned inversely to obtain a priori guidance. The prior losses of all layers are averaged to form a \underline{S}tatic \underline{K}nowledge \underline{D}istillation loss $L_{SKD}$. Finally, it is given a balance coefficient $\beta$ and combined with the cross-entropy $L_{CE}$ of the classification task to jointly optimize the student network.

\begin{equation}
\label{equ:skd}
\begin{aligned}
	 L_{SKD} &= - \frac{1}{L} \cdot \frac{1}{C} \sum_l^L \sum_{i,j}^C p(j|i) \log q(j|i) \\
	 		 &= - \frac{1}{L} \cdot \frac{1}{C} \sum_l^L \sum_{i,j}^C h(d(s^l_i, s^l_j)) \log h(d(a^l_i, a^l_i))
\end{aligned} 
\end{equation}

\begin{equation}
\label{equ:overall}
	 L = L_{CE} + \beta \cdot L_{SKD}
\end{equation}

\section{Experiments}

\subsection{Experiment Setup}

As the semantic information of neurons in neural networks is extracted as vectors, It is a potential way to enable direct comparison of neuron semantics, analysis, and benefit downstream tasks. Therefore, we design qualitative and quantitative experiments to assess the optimized vectors for the precise characterization of semantic relationships, the combined representation of semantics, and the transfer of extracted knowledge.

Interpretability experiments mainly focus on exploring the responses of the ResNet18 model pre-trained on ImageNet to different patterns on the Broden dataset~\cite{bau2017network}. The Broden dataset unifies 63305 color images of several densely labeled image data sets, which are of pixel-level semantic annotations. It contains examples of a broad range of objects, scenes, object parts, textures, and materials in a variety of contexts. The image size of the dataset is resized to $224 \times 224$. In addition, we also compare the results of static knowledge distillation and existing knowledge distillation methods across multiple architectures on the cifar100 dataset. The CIFAR-100 dataset~\cite{krizhevsky2009learning} contains 50,000 $32\times32$ color images in 100 different classes for training.

\begin{figure}[t]
	\centering
	\includegraphics[width=8cm,height=6.5cm]{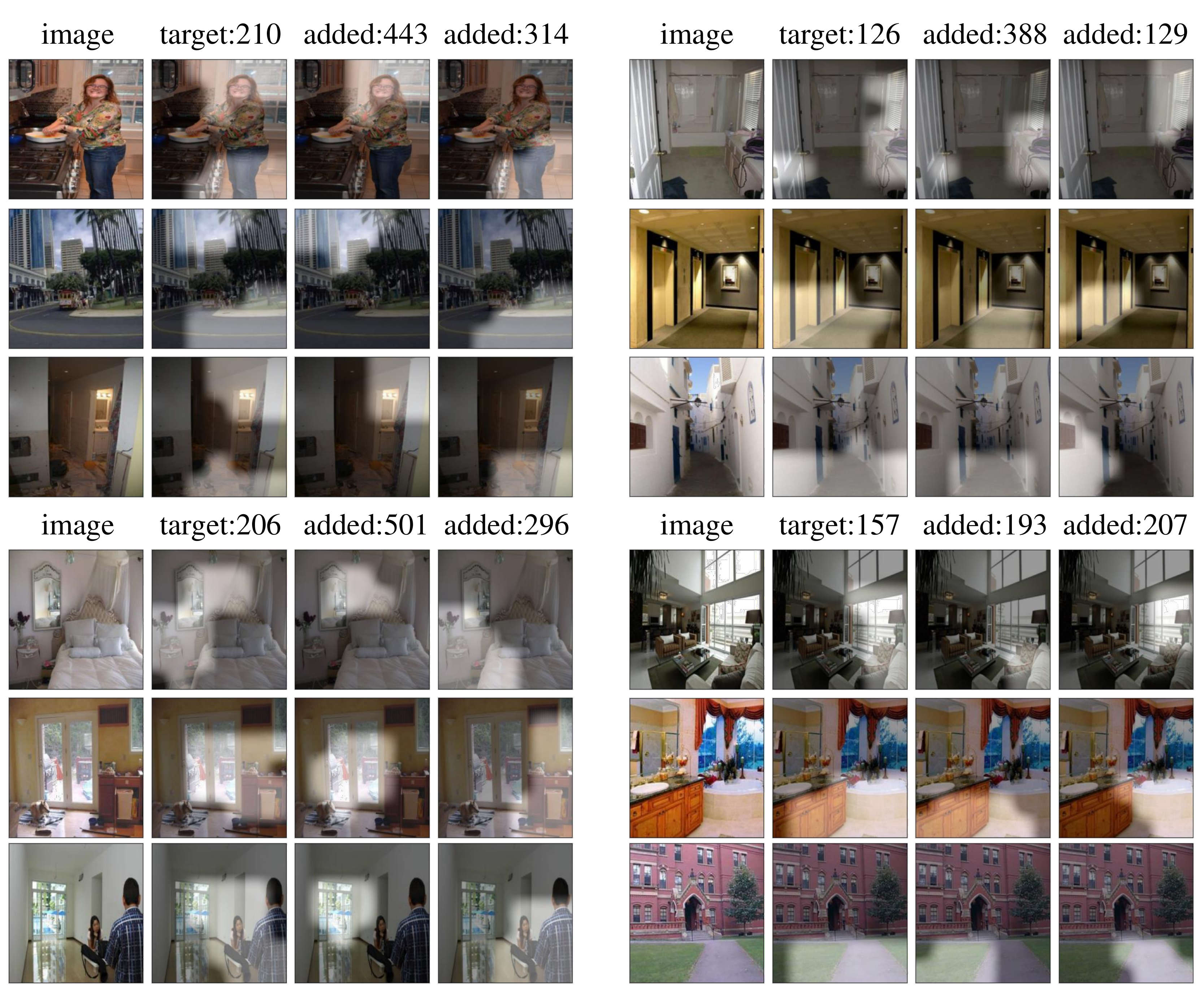}   
	\caption{Visualization of neuron additive relation. Through the guidance of semantic vectors, the complementary combination of neurons with simple semantics is discovered towards neurons with complex semantics} 
	\label{fig:add}
\end{figure}

\subsection{Implementation Details}
The semantic vector is extracted at the end of each block in neural networks, and its dimension is $16$. In all training, the batch size is $64$, and the weight decay is $5e^{-4}$. The SGD optimizer with an initial learning rate of $0.1$ is adopted. It is accompanied by a $0.1\times$ learning rate decay at \{50, 75, 90\} during 100 epochs. In the knowledge distillation, the balance coefficient $\beta$ for static knowledge distillation is set as $5e^{-1}$ for most architectures, and $5e^{-2}$ in the knowledge transfer from VGG13 to VGG8. The knowledge distillation experiments follow the protocol of CRD~\cite{tian2019contrastive}, and the results of comparison methods come from that paper. The results of our method are the mean of three trials.

\subsection{Qualitative visualization}

\subsubsection{Neuron Similar Relation}
The neuron semantic vector represents the correlation between neurons in the same layer. To intuitively assess whether such representation is accurate, we employ qualitative visualization of neuron similar relations. We demonstrate the activation maps of top-layer neurons and their three most similar neurons in different images, see Figure~\ref{fig:relationship}. The similarity between neurons is directly measured by the optimized semantic vectors. We do not limit the semantics of images. It can be seen that even in images with different semantics, the neurons with similar semantic vectors reflect similar activation. This proves that the semantic vector is a comprehensive representation of neuronal information.

\subsubsection{Neuron Additive Relation}
We exploit the additive relation between neuron semantic vectors, following Word2vec. The neuron semantic vectors are pairwise added. Then, the similarity is directly calculated between the combined semantic vector and the vector in the original space. Since the pre-trained model is not subject to interpretability constraints, some neurons show polysemy, as shown by target neurons in Figure~\ref{fig:add}. Since the semantic vector represents the comprehensive semantics of neurons, through its guidance, we can discover the complementary combination of neurons with simple semantics towards neurons with complex semantics.

\subsection{Quantitative metric}

\begin{figure}[t]
	\centering
	\includegraphics[width=6.7cm,height=4.9cm]{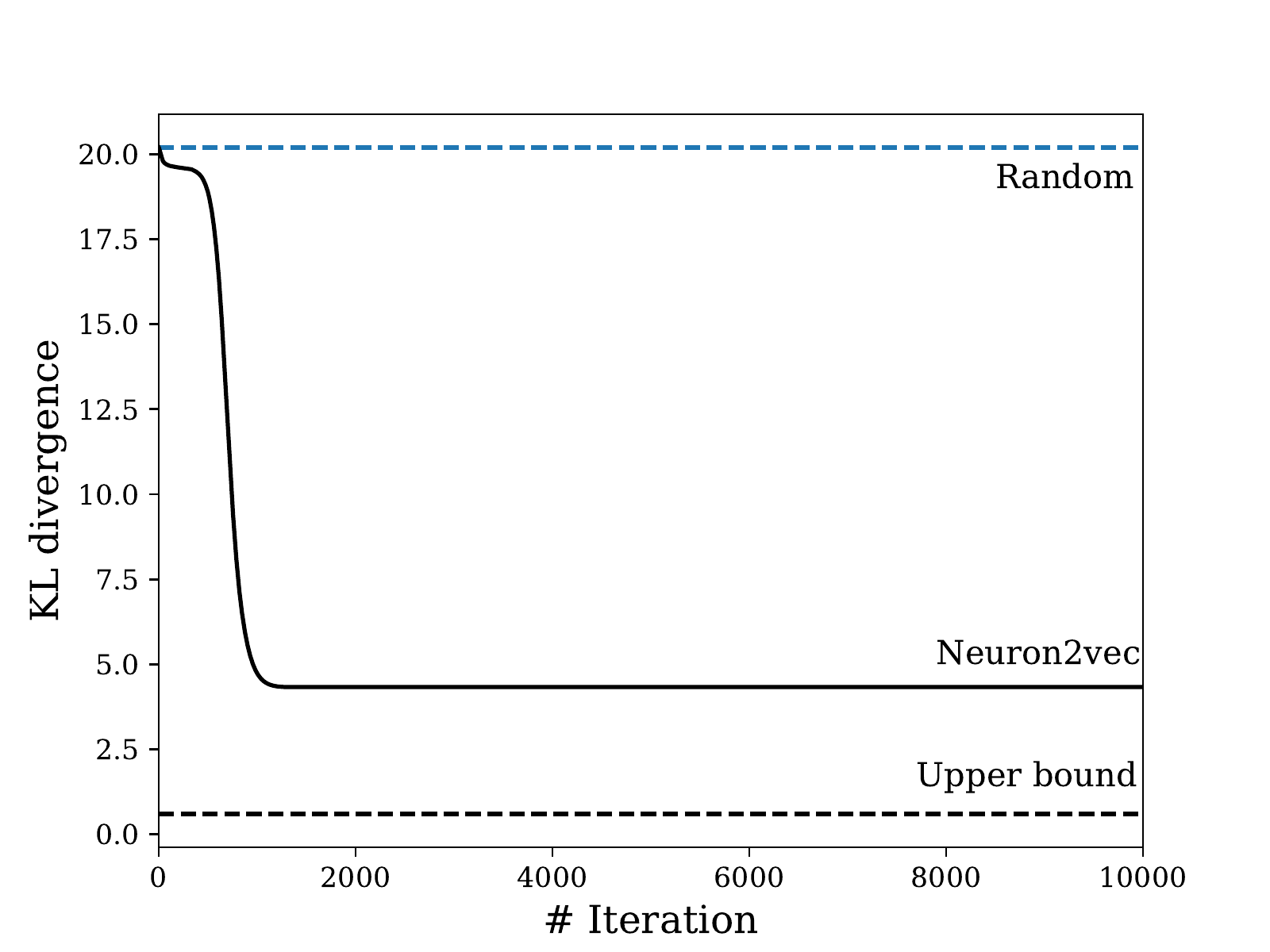}   
	\caption{KL curve during the training. Although the proposed neuron embedding method is unsupervised, the optimized semantic vectors are able to spontaneously converge to manually annotated semantic vectors.} 
	\label{fig:kl_curve}
\end{figure}

There is no ground truth for the neuron semantic vectors. However, NetDissect~\cite{bau2017network} assigns semantics to neurons based on their responses to the pixel-wise annotated image dataset. Inspired by that, we concatenate the response degree (i.e., intersection-over-union score) of the neuron to all semantics in the Broden dataset as the target of the neuron semantic vector. The quantitative metric of the semantic vector is measured by the KL divergence as the distribution distance between the optimized semantic vectors and the target semantic vectors. Figure~\ref{fig:kl_curve} illustrates the variation of the KL divergence during the training process. Since there is no directly relevant comparison algorithm, we only exhibit the result of random initialized semantic vectors as the baseline and the semantic vector optimized directly for such KL divergence as the upper bound. Without supervised information, the KL divergence curve in Figure~\ref{fig:kl_curve} decreases continuously during the training. It shows that the optimized semantic vector indeed encodes the semantic information of the neurons.

\subsection{Knowledge Distillation}
As knowledge distillation transfers knowledge from the large model to the small model, the improvement in classification accuracy of the small model could reflect the performance of knowledge distillation algorithms. As listed in Table~\ref{tab:kd}, we compare the proposed method with existing knowledge distillation algorithms in six types of cross-architectural knowledge migration tasks on the CIFAR100 dataset. The comparison methods include logit-based, feature-based and relation-based knowledge distillation algorithms, such as KD~\cite{hinton2015distilling}, PKT~\cite{passalis2018probabilistic}, CRD~\cite{tian2019contrastive}, FitNet~\cite{romero2014fitnets}, AT~\cite{zagoruyko2016paying}, VID~\cite{ahn2019variational}, AB~\cite{heo2019knowledge}, FT~\cite{kim2018paraphrasing}, NST~\cite{huang2017like}, FSP~\cite{yim2017gift}, SP~\cite{tung2019similarity}, CC~\cite{peng2019correlation}, and RKD~\cite{park2019relational}. The proposed static knowledge distillation is of a comparable or even better performance with existing relation-based knowledge distillation methods, without sample-by-sample guidance from the teacher model. It verifies the effectiveness of such an efficient and scalable knowledge distillation method.

\section{Conclusion}
In this paper, we propose a global interpretable method of neural networks, by embedding neurons into semantic space. With such static knowledge, we achieve decoupled two-stage knowledge distillation without sample-by-sample guidances from the teacher model. To this end, we align the relationship between neuron activation maps with the relationship between neuron semantic vectors. Such alignment is inversely applied from abstracted semantic vectors to neuron activation maps of the student model in the knowledge distillation. Qualitative and quantitative experiments have demonstrated that the proposed method describes neuron semantics well. Moreover, the static knowledge distillation shows its comparable or even superior performance compared with existing algorithms.

\acknowledgments{
The authors wish to thank A, B, and C. This work was supported in part by
a grant from XYZ (\# 12345-67890).}

\bibliographystyle{abbrv-doi}

\bibliography{template}

\begin{thebibliography}{10}

\bibitem{ahn2019variational}
S.~Ahn, S.~X. Hu, A.~Damianou, N.~D. Lawrence, and Z.~Dai.
\newblock Variational information distillation for knowledge transfer.
\newblock In {\em Proceedings of the IEEE/CVF Conference on Computer Vision and
  Pattern Recognition}, pp. 9163--9171, 2019.

\bibitem{bau2017network}
D.~Bau, B.~Zhou, A.~Khosla, A.~Oliva, and A.~Torralba.
\newblock Network dissection: Quantifying interpretability of deep visual
  representations.
\newblock In {\em Proceedings of the IEEE conference on computer vision and
  pattern recognition}, pp. 6541--6549, 2017.

\bibitem{bien2011prototype}
J.~Bien and R.~Tibshirani.
\newblock Prototype selection for interpretable classification.
\newblock {\em The Annals of Applied Statistics}, pp. 2403--2424, 2011.

\bibitem{caruana1999case}
R.~Caruana, H.~Kangarloo, J.~D. Dionisio, U.~Sinha, and D.~Johnson.
\newblock Case-based explanation of non-case-based learning methods.
\newblock In {\em Proceedings of the AMIA Symposium}, p. 212. American Medical
  Informatics Association, 1999.

\bibitem{crowley2018moonshine}
E.~J. Crowley, G.~Gray, and A.~J. Storkey.
\newblock Moonshine: Distilling with cheap convolutions.
\newblock {\em Advances in Neural Information Processing Systems}, 31, 2018.

\bibitem{fong2017interpretable}
R.~C. Fong and A.~Vedaldi.
\newblock Interpretable explanations of black boxes by meaningful perturbation.
\newblock In {\em Proceedings of the IEEE International Conference on Computer
  Vision}, pp. 3429--3437, 2017.

\bibitem{furlanello2018born}
T.~Furlanello, Z.~Lipton, M.~Tschannen, L.~Itti, and A.~Anandkumar.
\newblock Born again neural networks.
\newblock In {\em International Conference on Machine Learning}, pp.
  1607--1616. PMLR, 2018.

\bibitem{heo2019knowledge}
B.~Heo, M.~Lee, S.~Yun, and J.~Y. Choi.
\newblock Knowledge transfer via distillation of activation boundaries formed
  by hidden neurons.
\newblock In {\em Proceedings of the AAAI Conference on Artificial
  Intelligence}, vol.~33, pp. 3779--3787, 2019.

\bibitem{hinton2015distilling}
G.~Hinton, O.~Vinyals, J.~Dean, et~al.
\newblock Distilling the knowledge in a neural network.
\newblock {\em arXiv preprint arXiv:1503.02531}, 2(7), 2015.

\bibitem{huang2017like}
Z.~Huang and N.~Wang.
\newblock Like what you like: Knowledge distill via neuron selectivity
  transfer.
\newblock {\em arXiv preprint arXiv:1707.01219}, 2017.

\bibitem{kim2018paraphrasing}
J.~Kim, S.~Park, and N.~Kwak.
\newblock Paraphrasing complex network: Network compression via factor
  transfer.
\newblock {\em Advances in neural information processing systems}, 31, 2018.

\bibitem{krizhevsky2009learning}
A.~Krizhevsky, G.~Hinton, et~al.
\newblock Learning multiple layers of features from tiny images.
\newblock 2009.

\bibitem{mikolov2013distributed}
T.~Mikolov, I.~Sutskever, K.~Chen, G.~S. Corrado, and J.~Dean.
\newblock Distributed representations of words and phrases and their
  compositionality.
\newblock {\em Advances in neural information processing systems}, 26, 2013.

\bibitem{park2019relational}
W.~Park, D.~Kim, Y.~Lu, and M.~Cho.
\newblock Relational knowledge distillation.
\newblock In {\em Proceedings of the IEEE/CVF Conference on Computer Vision and
  Pattern Recognition}, pp. 3967--3976, 2019.

\bibitem{passalis2018probabilistic}
N.~Passalis and A.~Tefas.
\newblock Probabilistic knowledge transfer for deep representation learning.
\newblock {\em CoRR, abs/1803.10837}, 1(2):5, 2018.

\bibitem{peng2019correlation}
B.~Peng, X.~Jin, J.~Liu, D.~Li, Y.~Wu, Y.~Liu, S.~Zhou, and Z.~Zhang.
\newblock Correlation congruence for knowledge distillation.
\newblock In {\em Proceedings of the IEEE/CVF International Conference on
  Computer Vision}, pp. 5007--5016, 2019.

\bibitem{romero2014fitnets}
A.~Romero, N.~Ballas, S.~E. Kahou, A.~Chassang, C.~Gatta, and Y.~Bengio.
\newblock Fitnets: Hints for thin deep nets.
\newblock {\em arXiv preprint arXiv:1412.6550}, 2014.

\bibitem{shapley2016games}
L.~S. Shapley.
\newblock {\em A value for n-person games}.
\newblock Princeton University Press, 2016.

\bibitem{tian2019contrastive}
Y.~Tian, D.~Krishnan, and P.~Isola.
\newblock Contrastive representation distillation.
\newblock {\em arXiv preprint arXiv:1910.10699}, 2019.

\bibitem{tung2019similarity}
F.~Tung and G.~Mori.
\newblock Similarity-preserving knowledge distillation.
\newblock In {\em Proceedings of the IEEE/CVF International Conference on
  Computer Vision}, pp. 1365--1374, 2019.

\bibitem{yim2017gift}
J.~Yim, D.~Joo, J.~Bae, and J.~Kim.
\newblock A gift from knowledge distillation: Fast optimization, network
  minimization and transfer learning.
\newblock In {\em Proceedings of the IEEE Conference on Computer Vision and
  Pattern Recognition}, pp. 4133--4141, 2017.

\bibitem{zagoruyko2016paying}
S.~Zagoruyko and N.~Komodakis.
\newblock Paying more attention to attention: Improving the performance of
  convolutional neural networks via attention transfer.
\newblock {\em arXiv preprint arXiv:1612.03928}, 2016.

\bibitem{zhang2019your}
L.~Zhang, J.~Song, A.~Gao, J.~Chen, C.~Bao, and K.~Ma.
\newblock Be your own teacher: Improve the performance of convolutional neural
  networks via self distillation.
\newblock In {\em Proceedings of the IEEE/CVF International Conference on
  Computer Vision}, pp. 3713--3722, 2019.

\bibitem{zhou2016learning}
B.~Zhou, A.~Khosla, A.~Lapedriza, A.~Oliva, and A.~Torralba.
\newblock Learning deep features for discriminative localization.
\newblock In {\em Proceedings of the IEEE conference on computer vision and
  pattern recognition}, pp. 2921--2929, 2016.

\bibitem{zintgraf2017visualizing}
L.~M. Zintgraf, T.~S. Cohen, T.~Adel, and M.~Welling.
\newblock Visualizing deep neural network decisions: Prediction difference
  analysis.
\newblock {\em arXiv preprint arXiv:1702.04595}, 2017.

\end{thebibliography}
\end{document}